\documentclass[runningheads]{llncs}
\usepackage[T1]{fontenc}
\usepackage{graphicx}
\usepackage{comment}
\usepackage{subfig}
\usepackage[noadjust]{cite} 
\usepackage{xcolor} 
\usepackage{amssymb}
\usepackage{amsmath}
\usepackage{bbm}

\begin{document}
\title{Grouping Auction-Consensus Algorithm for Decentralized Task Allocation in Multi-Robot Systems}
\titlerunning{Grouping Auction-Consensus for Decentralized Task Allocation}
%

\author{Jose Rodriguez\inst{1}\orcidID{0009-0006-6161-0170} \and Sven Koenig\inst{2}\orcidID{0000-0002-5458-094X} \and Wenjie Dong\inst{1}\orcidID{0000-0003-2842-1782} \and Qi Lu\inst{1}\orcidID{0000-0002-8425-4686}}

\authorrunning{J. Rodriguez et al.}
%
\institute{The University of Texas at Rio Grande Valley, Edinburg, Texas 78539, USA \\  \email{\{jose.rodriguez53,wenjie.dong,qi.lu\}@utrgv.edu} 
 \and
University of California, Irvine, CA 92697, USA \\ \email{\{svenk\}@uci.edu}  
}
\maketitle              
\begin{abstract}
Decentralized multi-robot task allocation (MRTA) is essential for scalable and resilient autonomous systems. The Consensus-Based Bundle Algorithm (CBBA) is a widely adopted decentralized baseline. However, its individual task-level bidding is poorly aligned with the min-sum objective of minimizing total team travel distance, leading to suboptimal allocations in spatially distributed environments. This paper introduces the Grouping Auction-Consensus Algorithm (GACA). This decentralized MRTA framework adopts the two-phase auction-consensus architecture of CBBA while fundamentally redesigning its bidding mechanism to reason over groups of spatially proximate tasks. A nearest-neighbor preprocessing step partitions tasks into spatially coherent groups before allocation. Agents then iteratively propose
structured group-level actions: claiming unassigned groups, acquiring partial groups, or contesting groups held by other agents. Competing actions are resolved through a consensus phase. Operating in the MT-SR-IA problem class, GACA is evaluated against CBBA using a Mixed-Integer Linear Program as the ground-truth optimality reference. Across four swarm sizes and 4,000 test worlds, GACA achieves a median percent optimality of approximately 97\% compared to 81--84\% for CBBA, while converging in equal or fewer iterations. A scalability evaluation over 3,280 additional problem instances spanning swarm sizes of 5 to 20 agents and task counts of 10 to 50 confirms that these gains generalize robustly across a wide range of problem configurations.

\keywords{Multi-Robot Task Allocation \and Decentralized Auction-Consensus \and Min-Sum Optimization}
\end{abstract}
\section{Introduction}

The coordination of multiple autonomous agents to efficiently allocate and execute a set of tasks is a fundamental challenge in multi-robot systems. Multi-Robot Task Allocation (MRTA) problems arise in search and rescue, warehouse logistics, environmental monitoring, and UAV mission planning \cite{Swarm_Intelligence_Review}. Centralized approaches can achieve globally optimal solutions, but they suffer from single points of failure, poor scalability, and sensitivity to communication delays \cite{Swarm_Intelligence_Review}\cite{Opt_MRTA_Review}. Decentralized market-based auction algorithms have emerged as an effective alternative, enabling agents to self-organize through competitive bidding \cite{Market_Approach_Review}.

A landmark contribution is the Consensus-Based Bundle Algorithm (CBBA) \cite{cbba2009}, which alternates between a greedy bundle-building phase and a consensus phase that resolves conflicting task ownership through local communication. CBBA guarantees convergence to conflict-free solutions and is a widely adopted decentralized MRTA baseline.

Despite its success, CBBA treats tasks as independent entities. In many deployments, however, tasks form spatial clusters, and allocating them individually leads to inefficient routing and redundant agent travel. This paper introduces the Grouping Auction-Consensus Algorithm (GACA), a decentralized MRTA framework that adopts the two-phase auction-consensus architecture of CBBA while fundamentally redesigning its bidding mechanism to reason jointly over groups of spatially proximate tasks. Rather than bidding on individual tasks, GACA agents propose structured actions over task groups, where actions consist of claiming unassigned groups, acquiring partial groups, or contesting groups held by other agents. Competing proposals are then resolved through a consensus phase. GACA targets the min-sum objective of minimizing total team travel distance, and is designed to substantially outperform CBBA on this criterion.

The remainder of this paper is organized as follows. Section~2 reviews related work. Section~3 presents the problem formulation. Section~4 describes GACA in detail. Section~5 presents simulation results. Section~6 discusses the findings, and Section~7 concludes.

\section{Related Works}

\subsection{Multi-Robot Task Allocation}
The taxonomy of Gerkey and Matarić \cite{MRTA_Taxonomy} classifies MRTA problems along three axes: single- vs.\ multi-task robots (ST vs. MT), single- vs.\ multi-robot tasks (SR vs. MR), and instantaneous vs.\ time-extended assignment (IA vs. TA). This paper operates in the MT-SR-IA class. Robots execute multiple tasks, each task requires one robot, and assignments are made instantaneously. This problem definition has direct connections to the multiple Traveling Salesman Problem (mTSP) \cite{mTSP2006} and the min-sum criterion of minimizing total team travel distance.

\subsection{Consensus-Based Bundle Algorithm and Variants}
CBBA \cite{cbba2009} coordinates agents through decentralized bundle-building and consensus phases, guaranteeing convergence to a conflict-free assignment proven at least 50\% optimal for max-sum formulations. Performance guarantees for min-sum are considerably weaker, motivating the present work. Several extensions address specific limitations outside the MT-SR-IA scope considered in this paper. The Asynchronous CBBA (ACBBA) \cite{acbba2020} extended the original synchronized framework of CBBA to support real-time dynamic tasking environments. The Coupled-Constraint CBBA (CCBBA) \cite{ccbba2011} introduced pessimistic and optimistic bidding strategies to handle complex temporal and assignment interdependencies between tasks. Event-Driven CBBA (ED-CBBA) \cite{edcbba2025} addressed the communication overhead inherent to CBBA by introducing an event-based broadcasting strategy, achieving a 52\% reduction in communication requirements while maintaining equivalent task allocation performance. CBPA \cite{CBPA} extended the bidding mechanism to incorporate payload consumption as a resource constraint, enabling fault-tolerant and real-time scheduling in systems where robots carry heterogeneous payload capacities. ICBGA \cite{ICBGA} further augmented the consensus framework with time window and resource load constraints, supporting collaborative task allocation for UAV teams operating under tighter mission requirements. More recent work, such as AI-CBBA \cite{aicbba2024}, incorporated graph convolutional neural networks into the bundle-building phase to improve spatial reasoning and task coverage, demonstrating lower average distances traveled as the number of tasks scaled. While these contributions meaningfully extend CBBA's capabilities, they largely preserve the core structure of individual task-level bidding and do not fundamentally alter how agents reason about spatial task structure during allocation.

Regarding the min-sum objective specifically, Otte et al. \cite{Otte2020} conducted an analysis of auction-based algorithms across multiple objectives. The work showed that standard CBBA-style scoring functions are not inherently aligned with the min-sum objective, underscoring the need for allocation strategies better suited to this criterion.

\subsection{Clustering and Grouping Approaches}
Explicit task grouping before allocation has shown promise for min-sum objectives. Elango et al. \cite{K_Means_Auction} combined K-means clustering with an auction in a three-stage pipeline, reducing mTSP complexity through subproblem decomposition, though the cluster formation step is centralized. Yuan et al. \cite{K_means_PSO} combined K-means++ with Particle Swarm Optimization for cluster-to-robot assignment. In the decentralized setting, Dec-MATA \cite{dec-mata} formulates task planning as maximum-weighted bipartite matching with soft clustering, achieving solutions within 7–28\% of optimal MILP cost at orders-of-magnitude lower runtime. However, Dec-MATA's clustering is computed in a one-shot step without inter-agent negotiation, limiting its ability to adaptively split or merge groups as the allocation evolves.
Prior clustering-based MRTA methods share a common limitation: task groupings are determined either centrally or independently by each agent in a one-shot preprocessing step, without any mechanism for agents to iteratively contest, split, or merge groups as the allocation evolves. This restricts the flexibility of the allocation and can lead to suboptimal assignments when initial groupings are poorly matched to the spatial distribution of robots. GACA addresses this gap by embedding group-level negotiation directly into the auction-consensus loop, allowing agents to propose and resolve competing claims over task groups, including splitting groups and stealing assigned groups, across multiple iterations until convergence.

\section{Problem Formulation}

\subsection{Agents and Tasks}

Consider a team of $N_a$ homogeneous robots operating in a bounded two-dimensional workspace $\mathcal{W} \subset \mathbb{R}^2$. The set of agents is denoted $\mathcal{I} = \{1, 2, \ldots, N_a\}$, and each agent $i \in \mathcal{I}$ has an initial position $m_i \in \mathcal{W}$. The set of tasks to be completed is denoted $\mathcal{J} = \{1, 2, \ldots, N_t\}$, where each task $j \in \mathcal{J}$ is associated with a fixed location $n_j \in \mathcal{W}$ that an agent must physically visit to execute the task.

The Euclidean distance between an agent's initial position and a task location is written as:
\begin{equation}
    d_{m_i, n_j} = \|m_i - n_j\|_2
\end{equation}

and the Euclidean distance between two task locations is written as:

\begin{equation}
    d_{n_j, n_k} = \|n_j - n_k\|_2, \quad j \neq k
\end{equation}

\subsection{Communication Model}
The communication topology is an undirected graph $G_C=(\mathcal{I},\mathcal{E}_C)$ where $(i,k)\in\mathcal{E}_C$ if agents $i$ and $k$ are within direct communication range. The direct communication indicator
\begin{equation}
    g_{ik} =
    \begin{cases}
        1 & \text{if } (i,k) \in \mathcal{E}_C \\
        0 & \text{otherwise}
    \end{cases}
\end{equation}
denotes whether agents $i$ and $k$ can directly exchange information, with $g_{ik} = g_{ki}$ for all $i, k \in \mathcal{I}$ and $g_{ii} = 0$ by convention. During Phase~2, agent $i$ collects proposals from all $k$ with $g_{ik}=1$.

\subsection{Task Allocation as a Weighted Directed Graph}
The allocation is a weighted directed graph $G_A=(\mathcal{V},\mathcal{E},w)$ with $\mathcal{V}=\mathcal{I}\cup\mathcal{J}$ and $\mathcal{E}\subseteq(\mathcal{I}\times\mathcal{J})\cup(\mathcal{J}\times\mathcal{J})$ capturing agent-to-task and task-to-task transitions. Edge weights equal their respective Euclidean distances: $w(i,j)=d_{m_i,n_j}$ and $w(j,k)=d_{n_j,n_k}$. A route $r_i$ for agent $i$ is an ordered sequence of tasks drawn from $\mathcal{J}$, beginning at $m_i$ and visiting each assigned task in sequence. The route induces a path in $G_A$ consisting of one agent-to-task edge followed by zero or more task-to-task edges.

\subsection{MT-SR-IA Conditions and Objective}
Every task is visited by exactly one agent (SR condition):
\begin{equation}
  \sum_{v\in\mathcal{I}\cup\mathcal{J},\,v\neq j}\mathbbm{1}[(v,j)\in\mathcal{E}]=1,
  \quad\forall\,j\in\mathcal{J}
\end{equation}
where $\mathbbm{1}[\cdot]$ is an indicator function that equals a value of 1 if the enclosed condition is true and 0 otherwise.

Each vertex has at most one outgoing edge, naturally admitting the multi-task (MT) condition whereby an agent can leave from a task vertex to continue executing an ordered chain of tasks:
\begin{equation}
  \sum_{v\in\mathcal{J},\,v\neq u}\mathbbm{1}[(u,v)\in\mathcal{E}]\leq1,
  \quad\forall\,u\in\mathcal{V}
\end{equation}
The min-sum objective is to minimize the total team travel distance:
\begin{equation}
  G_A^*=\arg\min_{G_A}\sum_{(u,v)\in\mathcal{E}}w(u,v)
\end{equation}
This is equivalent to mTSP and is NP-hard in general \cite{MRTA_Taxonomy}. GACA pursues a high-quality decentralized approximation.

\section{Grouping Auction-Consensus Algorithm}

\subsection{Preprocessing: Spatial Task Grouping}
Because all agents are assumed to have perfect and complete observability of the task set $\mathcal{J}$, preprocessing requires no communication and is performed identically and independently by every agent before the first iteration.

\noindent \textbf{Step 1: Nearest-Neighbor Graph.}
Task groups are initialized by constructing an undirected weighted graph $G_G = (\mathcal{J}, \mathcal{E}_G)$, where each task $j \in \mathcal{J}$ creates an edge to its nearest neighboring task:

\begin{equation}
    e_j = \{j,\ k^*\}, \quad k^* = \arg\min_{k \in \mathcal{J},\ k \neq j}\ d_{n_j, n_k}
\end{equation}
Duplicate edges, or cases where two tasks independently identify each other as nearest neighbors, are retained as a single edge. Some vertices may initially accumulate more than two incident edges, requiring conflict resolution.

\noindent \textbf{Step 2: Conflict Resolution.}
A vertex with more than two incident edges conflicts with the route structure. Conflicts are resolved iteratively: the highest-weight edge incident to any conflict vertex is removed. If removal produces an isolated vertex, it is rerouted by adding a new edge to the nearest degree-one task. Rerouting produces at most one new edge per isolated task, so it cannot create a new conflict vertex or introduce a closed loop. This repeats until all vertices have degree of at most two, leaving a graph of simple path segments and closed loops.

\noindent \textbf{Step 3: Loop Opening.}
Closed loops can arise independently of conflict resolution in rare cases where tasks are mutually proximate in a cyclic pattern, such that each task's nearest neighbor forms a closed cycle. For each such loop, the highest-weight edge is removed, converting it to an open path with two distinct endpoints. After Steps~1--3, every connected component in $G_G$
is a simple path, and Step~4 proceeds on this conflict-free, loop-free graph.

\noindent \textbf{Step 4: Group Route Assignment.}
Each connected component of $G_G$ constitutes a task group $g_l$, an ordered bidirectional route. Unassigned groups carry a negative identifier $-l$ while a group claimed by agent $i\in\mathcal{I}$ carries identifier $i>0$. The initial group collection consists of unassigned groups $\mathcal{G}=\{g_{-1},\ldots,g_{-N_g}\}$.

\noindent \textbf{Agent-Local State Variables.}
Each agent $i$ maintains five lists throughout GACA. The \emph{route list} $p_i$ stores the bidirectional route believed for each group, where entry $p_{i,l}$ stores the bidirectional route that agent $i$ currently believes for group $l$. At initialization, $p_i = \mathcal{G}$ for all $i \in \mathcal{I}$. The \emph{group list} $z_i\in\mathbb{Z}^{N_t}$ records, for each task $j$, either its negative group identifier (unassigned) or the positive agent identifier that owns its group. At initialization, $z_{i,j}=l$ for all tasks $j\in g_l$. The \emph{bid list} $y_i\in\mathbb{R}^{N_t}_+$ stores the lowest observed bid per task. Lower bid values correspond to more competitive actions, so $y_{i,j}$ is initialized to $+\infty$ for all tasks, indicating that no bids have been placed. The \emph{ordering list} $o_i\in\mathbb{Z}_{\geq0}^{N_t}$ encodes each task's zero-indexed position within its group route. Together, $(z_i,o_i)$ fully encode the route structure $p_i$ and serve as the primary communication medium. The \emph{timestamp list} $s_i\in\mathbb{R}^{N_a}_+$ records the last received time from each agent, used during consensus to resolve stale information; initialized with $s_{i,i}$ at the current time and $s_{i,k}=0$ for $k\neq i$. Lists $z_i$, $o_i$, $y_i$, and $s_i$ are shared with neighbors during Phase 2.

\subsection{Phase 1: Action Selection}
Each agent $i$ independently proposes an action, or a desired change to the group allocation, encoded into its local $z_i$, $o_i$, $y_i$.
Route list $p_i$ is not modified until consensus confirms the action. Agent $i$ tracks its current route $p_{i,i}$ with route tail $\tau_i$, which defaults to $m_i$ if no tasks are yet assigned to agent $i$. Acquired groups are appended after $\tau_i$.

\noindent \textbf{Bid Structure.}
Unlike CBBA's per-task bids, GACA places one bid per action, and agents attempt at most one new action per action selection phase. Each action appends a group or sub-route to $p_{i,i}$ via a root task $r\in\mathcal{J}$ (the connection point). The bid $b$ encodes the total acquisition cost. Upon action selection, $y_{i,r}\leftarrow b$, and $z_i$, $o_i$ are updated for all affected tasks.

\noindent \textbf{Candidate Actions.}
Agent $i$ evaluates four action types and selects the one with lowest $b < y_{i,r}$:

\noindent \textit{Full Group Claim.} 
Claim an entire unassigned group $g_l$. The root $r$ is the endpoint of $g_l$ closer to $\tau_i$: $b=d_{\tau_i,n_r}$. All tasks in $g_l$ have $z_i$ updated from $-l$ to $i$, $o_i$ entries set to the appended ordering, and $y_{i,r}\leftarrow b$.

\noindent \textit{Split Claim.} 
Claim a contiguous sub-route of an unassigned $g_l$ from an interior root $r$ to one endpoint. The residual fragment must be appended to an unassigned group. $b=d_{\tau_i,n_r}+c_{\text{split}}$, where $c_{\text{split}}$ is the edge cost of severing the group at $r$ and rerouting the residual fragment with an unassigned group. Claimed tasks have $z_i$ updated to $i$, $o_i$ reindexed, and $y_{i,r}\leftarrow b$. Residual tasks receive a new negative identifier with $o_i$ reindexed accordingly.

\noindent \textit{Full Steal.} 
Claim an entire group currently owned by agent $k\neq i$, connecting at its nearer endpoint $r$. All tasks move from $k$ to $i$ in $z_i$, $o_i$ may be preserved or inverted, and $y_{i,r}\leftarrow b$.

\noindent \textit{Split Steal.} 
Claim a sub-route of agent $k$'s group from interior root $r$ to the last endpoint in agent $k$'s group. The tasks between $r$ and the groups connection to agent $k$ stay with $k$ while the rest of the tasks are moved to $i$. Split Steal uses identical cost structure and update rules to Split Claim.

If no action satisfies $b<y_{i,r}$, agent $i$ makes no proposal and leaves $z_i$, $o_i$, $y_i$ unchanged. The updated $(z_i,o_i,y_i,s_i)$ are passed to Phase~2.

\subsection{Phase 2: Consensus}
Each agent $i$ resolves conflicting proposals from all visible senders $\mathcal{S}_i=\{k\in\mathcal{I}:g_{ik}=1\}\cup\{i\}$, producing a revised $(z_i^*,y_i^*,o_i^*)$ for the next iteration.

\noindent \textbf{Proposal Collection and Action Set Extraction.}
For each sender $k\in\mathcal{S}_i$, agent $i$ reconstructs $k$'s proposed route list $\hat{p}_k$ by grouping tasks sharing a common $z_{k,j}$ value and sorting by $o_{k,j}$. Three sets are then extracted relative to agent $i$'s current state $z_i$. The \emph{claimed set} $C_k$ contains tasks $j$ where $z_{k,j}=k$ but $z_{i,j}\neq k$ — tasks $k$ is actively acquiring. The \emph{prerequisite set} $R_k$ contains tasks where both $z_{k,j}=k$ and $z_{i,j}=k$ — tasks already owned by $k$ whose continued ownership is required for $k$'s proposed new actions to be valid. The \emph{donation map} $D_k$ contains tasks with $z_{k,j}<0$ and $z_{k,j}\neq z_{i,j}$ — tasks being released to unassigned status. This occurs either when sender $k$ performs a split action (releasing the residual fragment) or when a prior competing claim outcompeted $k$, invalidating downstream tasks that must be returned to the unassigned pool.

\noindent \textbf{Conflict Graph Construction.}
Agent $i$ builds an undirected conflict graph $\mathcal{F}=(\mathcal{S}_i,\mathcal{E}_\mathcal{F})$ by connecting any pair of senders $(a,b)$ that exhibit at least one of four incompatibilities: a \emph{claim conflict} ($C_a\cap C_b\neq\emptyset$), a \emph{prerequisite conflict} ($C_a\cap R_b\neq\emptyset$ or vice versa), a \emph{donation conflict} (one sender donates a task while the other claims from the same group), or an \emph{ordering conflict} (both reference the same group with disagreeing $(z,o)$ entries). Agent $i$ then decomposes $\mathcal{F}$ into connected components $\{\mathcal{K}_1,\ldots,\mathcal{K}_M\}$ via depth-first search; senders in different components are resolved independently.

\noindent \textbf{Winner Selection and Application.}
For each component $\mathcal{K}_m$ with task set $T_m=\bigcup_{k\in\mathcal{K}_m}C_k$, agent $i$ computes the \emph{total allocation cost}
\begin{equation}
  \Phi_k=\sum_{l\in\mathcal{L}_k}\bigl[\operatorname{cost}(\hat{p}_{k,l})+y_{k,r_l}\bigr]
\end{equation}
for each sender $k$, where $\mathcal{L}_k$ is the index set of all groups in $\hat{p}_k$, $\operatorname{cost}(\hat{p}_{k,l})$ is total inter-task distance for group $l$, and $y_{k,r_l}$ is the bid for group $l$'s entry task. Scoring over the full route list $\hat{p}_k$ rather than only contested tasks ensures equitable comparison regardless of prior task accumulation. If $T_m=\emptyset$ (pure ordering disagreement), the lowest-$\Phi_k$ sender wins and its full route list is applied coherently. When $T_m\neq\emptyset$, the winner $k^*=\arg\min_{k\in\mathcal{K}_m}\Phi_k$ among active claimants is selected. Agent $i$ applies $k^*$'s allocation over all affected tasks, updating $z_i^*[j]\leftarrow z_{k^*,j}$ and $o_i^*[j]\leftarrow o_{k^*,j}$ for every task in the affected group set $\mathcal{A}_{k^*}$. This extension to all tasks in affected groups, not only explicitly claimed ones, prevents stale $z$ and $o$ entries from producing incoherent routes in subsequent iterations. Donation tasks from $D_{k^*}$ are applied to their target negative identifiers, with existing task positions reindexed accordingly.

After all components are processed, agent $i$ performs a final pass to propagate uncontested ownership. For tasks not covered by any $C_k$ whose group has not been resolved, if all senders in $\mathcal{S}_i$ that mention task $j$ agree on the same $z$ and $o$ values, those values are written into $z_i^*[j]$ and $o_i^*[j]$. Route list $p_i^*$ is then reconstructed by grouping and sorting, and $y_i^*$ is updated by inheriting bid values from the winning sender for any task that changed ownership.

\noindent \textbf{Convergence.}
GACA iterates between Phase~1 and Phase~2 until no agent proposes a new action and all senders agree on a conflict-free allocation.

\section{Simulative Results}

\subsection{Simulation Setup}
\begin{figure}[!htbp]
 \centering
  \includegraphics[width=0.6\textwidth]{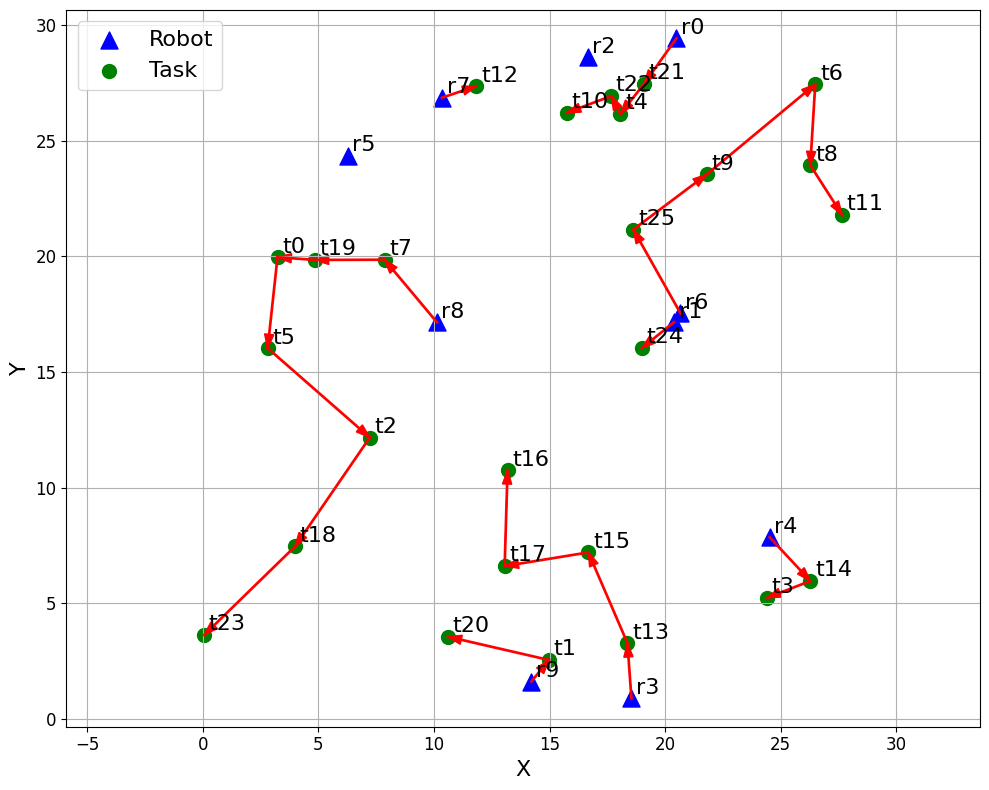}
  \caption{The optimal solution for 10 robots and 25 tasks in simulation}
  \label{example_solution}
\end{figure}

Simulations were conducted on randomly generated worlds with agent and task positions uniformly sampled in square workspaces, with random side lengths between 25–55 units. Task counts ranged from 2 to 4 per agent, yielding 10–20 tasks for 5-agent teams and 40–80 tasks for 20-agent teams. Four swarm sizes of 5, 10, 15, and 20 agents were evaluated over 1000 independently sampled validation worlds for each tested swarm size. All simulations used a fully connected communication graph ($g_{ik}=1$ for all $i\neq k$). Ground-truth optimal solutions were obtained via a mixed-integer linear programming (MILP) formulation with PuLP \cite{pulp2011}. 
Because MILP solving is NP-hard, computation time grows rapidly with problem size, limiting the scalability of MILP-optimal ground-truth solutions. As a result, extensive testing was limited to worlds with up to 20 agents. Note that MILP solutions serve only as a reference and are not required by the proposed GACA approach. An example optimal solution obtained via the MILP formulation is shown in Fig.~\ref{example_solution}. Note that in Fig.~\ref{example_solution}, some agents remain idle. This is an expected outcome, as the focus of this work is to approach optimal solutions for the MRTA min-sum objective. Balancing this objective against a secondary goal of minimizing agent idleness is outside the scope of this work and is left for future work.
Two metrics were recorded: \emph{percent optimality} (ratio of MILP cost to algorithm cost, as a percentage) and \emph{convergence iterations}. Demonstration videos are available on our YouTube playlist\footnote{\url{https://tinyurl.com/DARSTaskAllocation}}.

\subsection{Solution Quality}
\begin{figure}[!htbp]
 \centering
  \includegraphics[width=0.7\textwidth]{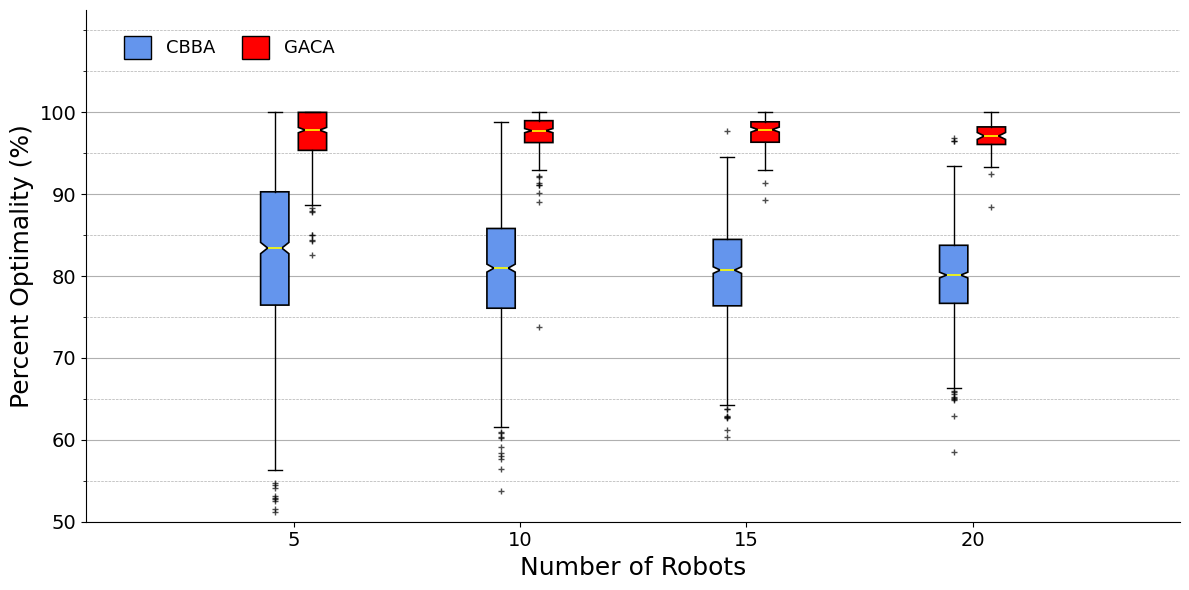}
  \caption{Percent optimality comparison between CBBA and GACA at
  varying robot team sizes.}
  \label{percent_optimality}
\end{figure}

Fig.~\ref{percent_optimality} shows percent optimality distributions for CBBA and GACA across all four swarm sizes. GACA achieves a median of approximately 97\% across all conditions, with an interquartile range of roughly 95–99\%. CBBA achieves a median of 81–84\% with a considerably wider spread, and the interquartile range spans 77–90\%, with the lower whisker falling below 65\% in several configurations. Notably, both algorithms produce tighter distributions as swarm size increases, but GACA's distribution remains consistently tighter than CBBA's at every swarm size, indicating that GACA's solution quality is not only higher but also more predictable as the problem scales.

\begin{figure}[!htbp]
 \centering
  \includegraphics[width=0.75\textwidth]{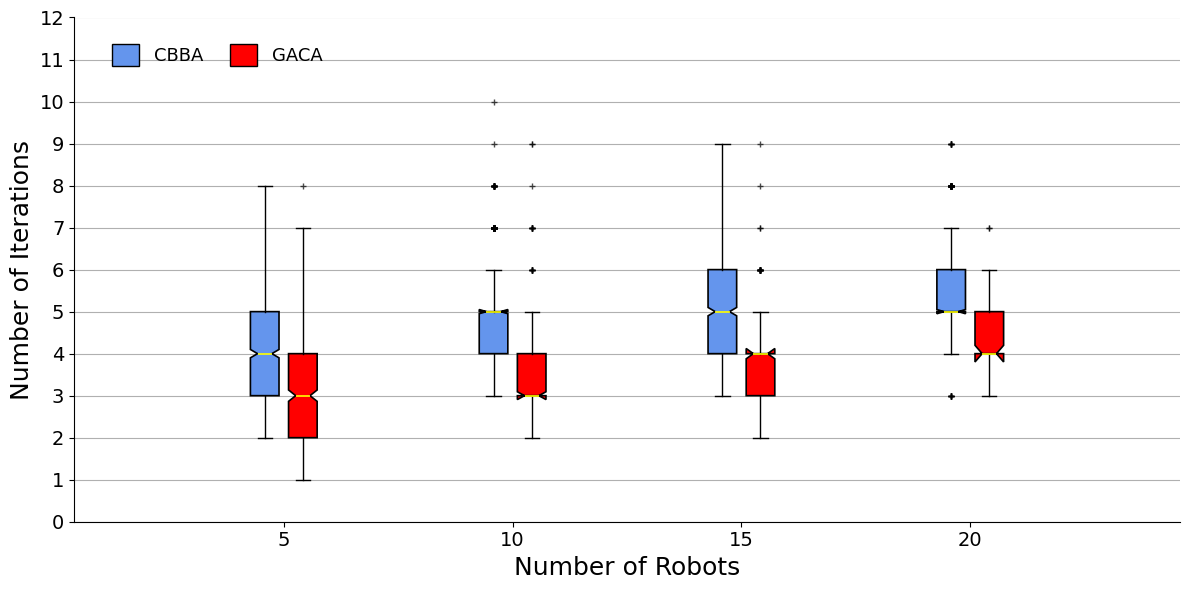}
  \caption{Convergence speed comparison between CBBA and GACA at varying
  robot team sizes (communication cycles until convergence).}
  \label{number_iterations}
\end{figure}

Fig.~\ref{number_iterations} shows convergence iteration distributions. GACA converges in equal or fewer iterations than CBBA across all swarm sizes, with a median approximately 1 iteration lower than CBBA.

\subsection{Scalability Evaluation Across Diverse Problem Instances}
\begin{figure}[!htbp]
 \centering
  \includegraphics[width=0.7\textwidth]{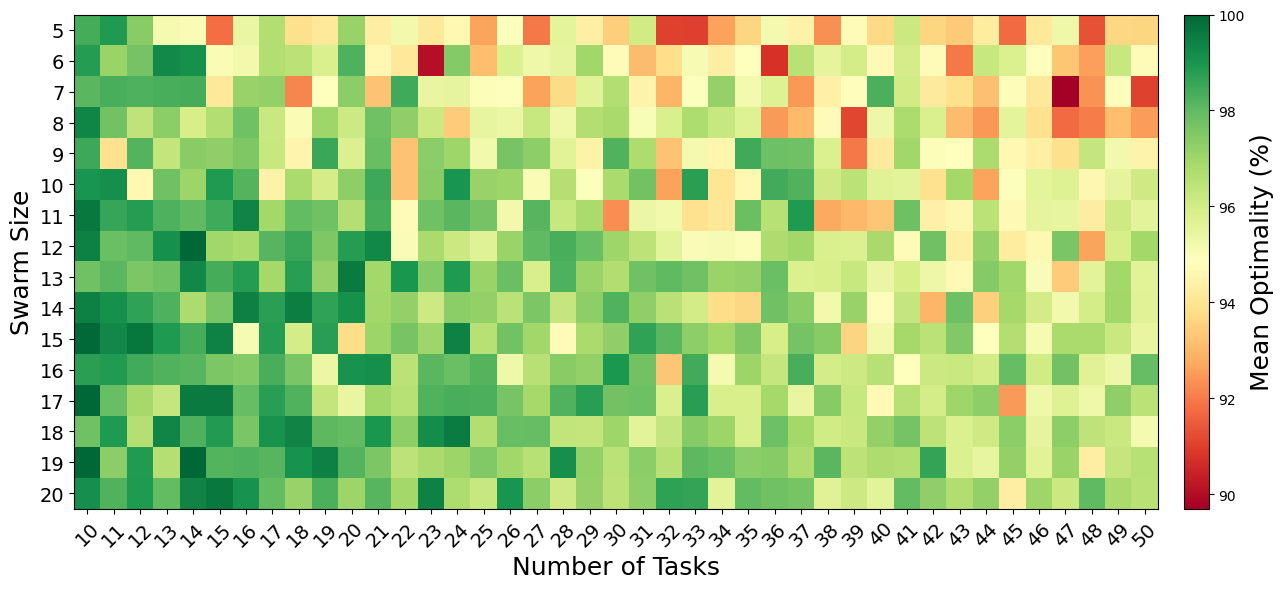}
  \caption{Mean percent optimality across the variety dataset, averaged
  over five world configurations per cell. Greener cells indicate higher
  optimality relative to the MILP ground truth.}
  \label{fig:heatmap_optimality}
\end{figure}

\begin{figure}[!htbp]
 \centering
  \includegraphics[width=0.7\textwidth]{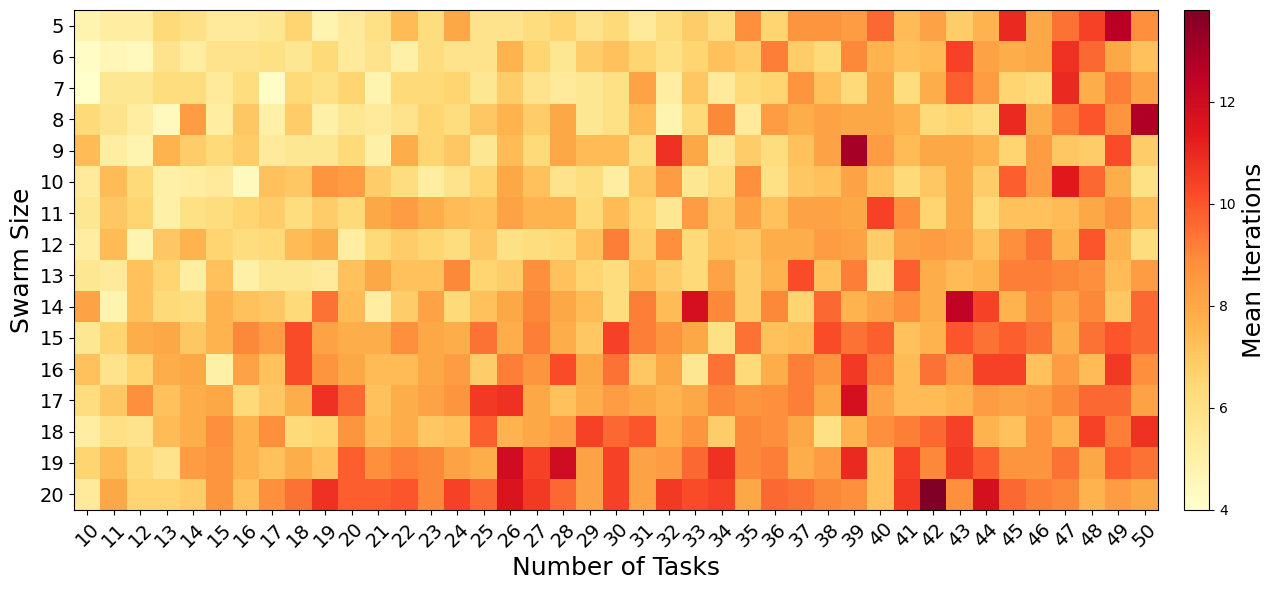}
  \caption{Mean convergence iterations across the variety dataset,
  averaged over five world configurations per cell. Darker red indicates
  higher iteration counts.}
  \label{fig:heatmap_iters}
\end{figure}

To characterize robustness beyond four fixed swarm sizes, a variety dataset was constructed spanning all integer swarm sizes from 5 to 20 agents and all integer task counts from 10 to 50, with five independently sampled world configurations per combination for 3,280 total instances. Mean percent optimality and mean convergence iterations were aggregated per (swarm size, task count) cell and displayed as heatmaps in Figs.~\ref{fig:heatmap_optimality} and~\ref{fig:heatmap_iters}.

Fig.~\ref{fig:heatmap_optimality} reveals a diagonal gradient: cells in the lower-left region, where swarm size is large relative to task count, tend toward darker green, while the upper-right region, where fewer agents cover proportionally more tasks, shows more yellow and orange shading. This suggests solution quality is influenced by the task-to-agent ratio. Fig.~\ref{fig:heatmap_iters} shows a complementary pattern: lighter yellow cells concentrate in the upper-left where swarm sizes and task counts are both small, while darker red cells appear more frequently in the lower-right as larger swarms contend over more tasks. Despite these trends, performance differences across the parameter space are moderate rather than severe. All cells remain above approximately 90\% mean optimality, and all configurations converge within 14 iterations in the worst observed cases.

\section{Discussion}

The 13–16 percentage point median optimality advantage of GACA over CBBA stems directly from the alignment between GACA's bid structure and the min-sum objective. CBBA's task-level greedy bids optimize individual marginal rewards without accounting for the spatial coherence of full routes. GACA bids on spatially contiguous groups whose intra-group travel is minimized by construction, with a bid cost anchored to the entry distance from the agent's route tail — a signal directly related to the incremental team travel cost. This structural alignment enables near-optimal solutions where CBBA consistently falls short.

The observation that GACA converges in fewer iterations despite a richer action space reflects the atomic nature of group-level transfers: a single Full Group Claim or Full Steal simultaneously resolves the ownership of all tasks in a group, whereas CBBA may perform several contested cycles for individual tasks. Group-level actions also compress variance in convergence speed by resolving spatially coherent blocks rather than isolated tasks, explaining GACA's tighter iteration distribution. 
The variety dataset confirms that both properties hold broadly across 656 parameter combinations. A higher task-to-agent ratio produces moderate reductions in solution quality and moderate increases in convergence iterations, consistent with greater inter-agent contention over groups, but all tested configurations remain above roughly 90\% optimality and converge within 14 iterations.

\section{Conclusion}

This paper introduced GACA, a decentralized MRTA framework that adopts CBBA's two-phase auction-consensus architecture while replacing its individual task-level bidding with a group-level action space comprising four proposal types: Full Group Claim, Split Claim, Full Steal, and Split Steal. Evaluated against CBBA and a MILP ground truth, GACA achieved a median percent optimality of approximately 97\% versus 81–84\% for CBBA across four swarm sizes and 4000 total test worlds, while converging in equal or fewer iterations. A broad scalability evaluation over 3,280 instances confirmed that high solution quality and stable convergence generalize across a wide range of team sizes and task counts. These results demonstrate that embedding group-level negotiation into the auction-consensus loop is an effective and robust mechanism for near-optimal decentralized min-sum task allocation.

\noindent \textbf{Limitations and Future Work}.
The nearest-neighbor preprocessing is agnostic to agent starting positions, potentially producing group boundaries poorly matched to the robot deployment. Incorporating agent geometry into the grouping step is a natural improvement. GACA currently assumes homogeneous agents with no capacity constraints, time windows, or capability differences. Extending the bid structure and consensus rules to heterogeneous settings would broaden applicability. GACA is adaptable to function with more diverse communication topologies or time-varying communication graphs by appropriately using the timestamp lists $s_i$ to compensate for communication discrepancies, but for simplicity and to first provide proof of concept, the results shown in this work currently cover only evaluation with a fully connected communication graph. As such, while the core algorithm performs as expected under these idealized, time-invariant conditions, its robustness to adverse or imperfect communication has not been extensively tested. Since real-world deployments commonly experience imperfect communication, evaluating the robustness of GACA under a variety of communication conditions is an important next step. Finally, benchmarking against Dec-MATA and centralized clustering pipelines, and extending to 3D or dynamic task sets, are important directions.

\begin{credits}
\subsubsection{\ackname} 
The authors acknowledge the financial support provided by the NSF CISE Expand AI program (No. 2434916), the NSF CREST Center for Multidisciplinary Research Excellence in Cyber-Physical Infrastructure Systems (MECIS) (No. 2112650), and the NSF CISE MSI programs (No. 2318682 \& No. 2431569). We acknowledge the valuable discussion with Dr. Pascal Van Hentenryck from the NSF AI Center, AI4OPT, about the decentralized task allocation framework.   


\end{credits}

%
%
%
\bibliography{references}
\bibliographystyle{splncs04}

%





\end{document}